\documentclass[graybox]{svmult}

\usepackage{mathptmx}       
\usepackage{helvet}         
\usepackage{courier}        
\usepackage{type1cm}        
\usepackage{makeidx}         
\usepackage{graphicx}        
\usepackage{multicol}        
\usepackage[bottom]{footmisc}

\makeindex             

\begin{document}

\title*{Enhanced Optimization with Composite Objectives and Novelty Pulsation}
\author{Hormoz Shahrzad, Babak Hodjat, Camille Dollé, Andrei Denissov, Simon Lau, Donn Goodhew, Justin Dyer, Risto Miikkulainen}
\authorrunning{Hormoz Shahrzad et al.}
\institute{Hormoz Shahrzad, Babak Hodjat 
\at Cognizant Technology Solutions \\
\email{\{hormoz, babak\}@cognizant.com}
\and Camille Dollé, Andrei Denissov, Simon Lau, Donn Goodhew, Justin Dyer 
\at Sentient Investment Management \\
\email{\{camille.dolle, andrei.denissov, simon.lau, donn, justin.dyer\}@sentientim.com}
\and Risto Miikkulainen 
\at Cognizant Technology Solutions, 
\email{risto@cognizant.com}
\at The University of Texas at Austin,
\email{risto@cs.utexas.edu}}
\maketitle

\abstract{An important benefit of multi-objective search is that it maintains a diverse population of candidates, which helps in deceptive problems in particular. Not all diversity is useful, however: candidates that optimize only one objective while ignoring others are rarely helpful. A recent solution is to replace the original objectives by their linear combinations, thus focusing the search on the most useful tradeoffs between objectives. To compensate for the loss of diversity, this transformation is accompanied by a selection mechanism that favors novelty. This paper improves this approach further by introducing novelty pulsation, i.e. a systematic method to alternate between novelty selection and local optimization.  In the highly deceptive problem of discovering minimal sorting networks, it finds state-of-the-art solutions significantly faster than before. In fact, our method so far has established a new world record for the 20-line sorting network with 91 comparators. In the real-world problem of stock trading, it discovers solutions that generalize significantly better on unseen data. Composite Novelty Pulsation is therefore a promising approach to solving deceptive real-world problems through multi-objective optimization.}

\section{Introduction}
\label{sec:1}
Multi-objective optimization is most commonly used for discovering a Pareto front from which solutions that represent useful tradeoffs between objectives can be selected \cite{ref-5,ref-8,ref-9, ref-10, ref-15}. Evolutionary methods are a natural fit for such problems because the Pareto front naturally emerges in the population maintained in these methods. Interestingly, multi-objectivity can also improve evolutionary optimization because it encourages populations with more diversity. Even when the focus of optimization is to find good solutions along a primary performance metric, it is useful to create secondary dimensions that reward solutions that are different in terms of structure, size, cost, consistency, etc. Multi-objective optimization then discovers stepping stones that can be combined to achieve higher fitness along the primary dimension \cite{ref-26}. The stepping stones are useful in particular in problems where the fitness landscape is deceptive, i.e. where the optima are surrounded by inferior solutions \cite{ref-20}.

However, not all such diversity is helpful. In particular, candidates that optimize one objective only and ignore the others are less likely to lead to useful tradeoffs, and they are less likely to escape deception. Prior research demonstrated that it is beneficial to replace the objectives with their linear combinations, thus focusing the search in more useful areas of the search space, and make up for the lost diversity by including a novelty metric in parent selection \cite{ref-31}. This paper improves upon this approach by introducing the concept of novelty pulsation: the novelty selection is turned on and off periodically, thereby allowing exploration and exploitation to leverage each other repeatedly.

This idea is tested in two domains. The first one is the highly deceptive domain of sorting networks \cite{ref-17} used in the original work on composite novelty selection \cite{ref-31}. Such networks consist of comparators that map any set of numbers represented in their input lines to a sorted order in their output lines. These networks have to be correct, i.e. sort all possible cases of input. The goal is to discover networks that are as small as possible, i.e. have as few comparators organized in as few sequential layers as possible. While correctness is the primary objective, it is actually not that difficult to achieve, because it is not deceptive. Minimality, on the other hand, is highly deceptive and makes the sorting network design an interesting benchmark problem. The experiments in this paper show that while the original composite novelty selection and its novelty-pulsation-enhanced version both find state-of-the-art networks up to 20 input lines, novelty pulsation finds them significantly faster. It also beat the state of the art for 20-line network by finding a 91 comparators design, which broke the previous world record of 92 \cite{ref-32}.

The second domain is the highly challenging real-world problem of stock trading. The goal is to evolve agents that decide whether to buy, hold, or sell particular stocks over time in order to maximize returns. Compared to original composite novelty method, novelty pulsation finds solutions that generalize significantly better to unseen data. It therefore forms a promising foundation for solving deceptive real-world problems through multi-objective optimization.

\section{Background and Related Work}
\label{sec:2}
Evolutionary methods for optimizing single-objective and multi-objective problems are reviewed, as well as the idea of using novelty to encourage diversity and the concept of exploration versus exploitation in optimization methods. The domains of minimal sorting networks and automated stock trading are introduced and prior work in them reviewed.

\subsection{Single-Objective Optimization}
\label{subsec:1}
When the optimization problem has a smooth and non-deceptive search space, evolutionary optimization of a single objective is usually convenient and effective. However, we are increasingly faced with problems of more than one objective and with a rugged and deceptive search space. The first approach often is to combine the objectives into a single composite calculation \cite{ref-8}:

\begin{equation}
\label{eq-1}
\\{Composite}\left( O_{1},\ O_{2},\ldots,O_{k} \right) = \sum_{i = 1}^{k}{\alpha_{i}O_{i}^{\beta_{i}}}\;
\end{equation}

Where the constant hyper-parameters \(\alpha_{i}\) and \(\beta_{i}\) determine the relative importance of each objective in the composition. The composite objective can be parameterized in two ways:

\begin{enumerate}
\item{By folding the objective space, and thereby causing a multitude of solutions to have the same value. Diversity is lost since solutions with different behavior are considered to be equal.}
\item{By creating a hierarchy in the objective space, and thereby causing some objectives to have more impact than many of the other objectives combined. The search will thus optimize the most important objectives first, which in deceptive domains might result in inefficient search or premature convergence to local optima.}
\end{enumerate}

Both of these problems can be avoided by casting the composition explicitly in terms of multi-objective optimization.

\subsection{Multi-Objective Optimization}
\label{subsec:2}
Multi-objective optimization methods construct a Pareto set of solutions \cite{ref-10}, and therefore eliminate the issues with objective folding and hierarchy noted in Section \ref{subsec:1}. However, not all diversity in the Pareto space is useful. Candidates that optimize one objective only and ignore the others are less likely to lead to useful tradeoffs, and are less likely to escape deception.

One potential solution is reference-point based multi-objective methods such as NSGA-III \cite{ref-9, ref-10}. They make it possible to harvest the tradeoffs between many objectives and can therefore be used to select for useful diversity as well, although they are not as clearly suited for escaping deception.

Another problem with purely multi-objective search is crowding. In crowding, objectives that are easier to explore end up with disproportionately dense representation on the Pareto front. NSGA II addresses this problem by using the concept of crowding distance \cite{ref-8}, and NSGA III improves upon it using reference points \cite{ref-9, ref-10}. These methods, while increasing diversity in the fitness space, do not necessarily result in diversity in the behavior space.

An alternative method is to use composite multi-objective axes to focus the search on the area with most useful tradeoffs \cite{ref-31}. Since the axes are not orthogonal, solutions that optimize only one objective will not be on the Pareto front. The focus effect, i.e. the angle between the objectives, can be tuned by varying the coefficients of the composite.

However, focusing the search in this manner has the inevitable side effect of reducing diversity. Therefore, it is important that the search method makes use of whatever diversity exists in the focused space. One way to achieve this goal is to incorporate a preference for novelty into selection.

\subsection{Novelty Search}
\label{subsec:3}
Novelty search \cite{ref-23, ref-25} is an increasingly popular paradigm that overcomes deception by ranking solutions based on how different they are from others. Novelty is computed in the space of behaviors, i.e., vectors containing semantic information about how a solution performs during evaluation. However, with a large space of possible behaviors, novelty search can become increasingly unfocused, spending most of its resources in regions that will never lead to promising solutions.

Recently, several approaches have been proposed to combine novelty with a more traditional fitness objective \cite{ref-11, ref-13, ref-29, ref-30} to reorient search towards fitness as it explores the behavior space. These approaches have helped scale novelty search to more complex environments, including an array of control \cite{ref-2, ref-7, ref-29} and content generation \cite{ref-19, ref-21, ref-22} domains.

Many of these approaches combine a fitness objective with a novelty objective in some way, for instance as a weighted sum \cite{ref-6}, or as different objectives in a multi-objective search \cite{ref-29}. Another approach is to keep the two kinds of search separate, and make them interact through time. For instance, it is possible to first create a diverse pool of solutions using novelty search, presumably overcoming deception that way, and then find solutions through fitness-based search \cite{ref-18}. A third approach is to run fitness-based search with a large number of objective functions that span the space of solutions, and use novelty search to encourage search to utilize all those functions \cite{ref-7, ref-28, ref-30}. A fourth category of approaches is to run novelty search as the primary mechanism, and use fitness to select among the solutions. For instance, it is possible to add local competition through fitness to novelty search \cite{ref-22, ref-23}. Another version is to accept novel solutions only if they satisfy minimal performance criteria \cite{ref-11, ref-24}. Some of these approaches have been generalized using the idea of behavior domination to discover stepping stones \cite{ref-26, ref-27}.

In the Composite Novelty method \cite{ref-31}, a novelty measure is employed to select which individuals to reproduce and which to discard. In this manner, it is integrated into the genetic algorithm itself, and its role is to make sure the focused space that the composite multiple objectives define is searched thoroughly.

\subsection{Exploration versus Exploitation}
\label{subsec:4}
Every search algorithm needs to both explore the search space and exploit the known good solutions in it. Exploration is the process of visiting entirely new regions of a search space, whilst exploitation is the process of visiting regions within the neighborhood of previously visited points. In order to be successful, a search algorithm needs to establish a productive synergy between exploration and exploitation \cite{ref-35}.

A common problem in evolutionary search is that it gets stuck in local minima, i.e. in unproductive exploitation. A common solution is to kick-start the search process in such cases by temporarily increasing mutation rates. This solution can be utilized more systematically by making such kick-starts periodic, resulting in methods such as in delta coding and burst mutation \cite{ref-33, ref-34}.

This paper incorporates the kick-start idea into novelty selection. By turning novelty selection on and off periodically allows local search (i.e. exploitation) and novelty search (i.e. exploration) to leverage each other, leading to faster search and better generalization. These effects will be demonstrated in the sorting networks and stock trading domains, respectively.

\begin{figure}
\begin{center}
\includegraphics[scale=.15]{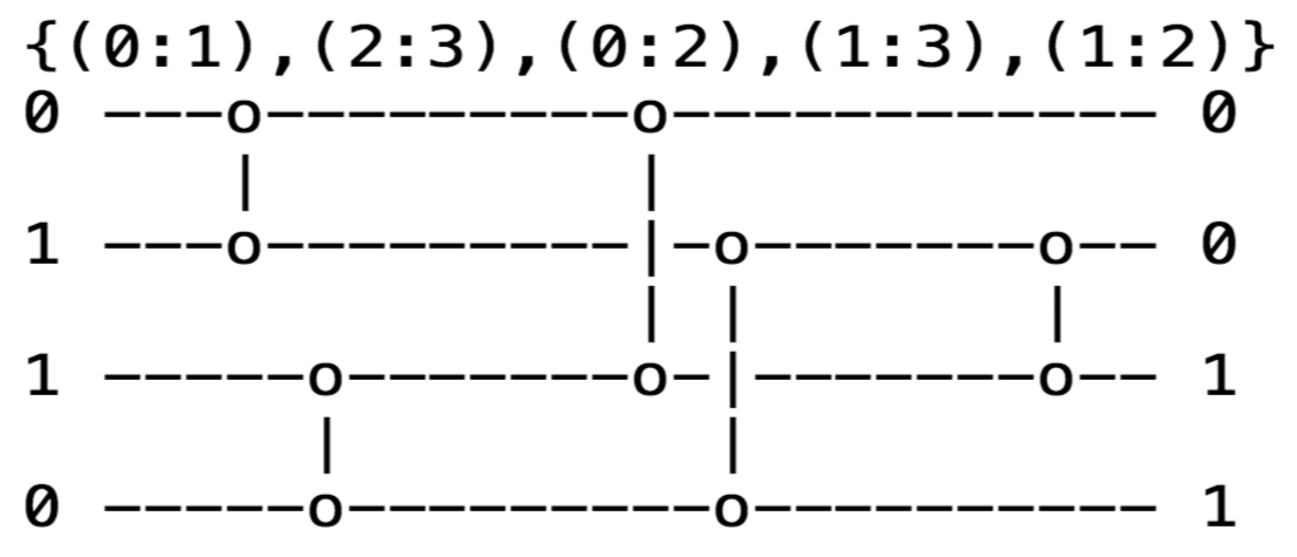}
\end{center}
\caption{A Four-Input Sorting Network and its representation. This network takes as its input (left) four numbers, and produces output (right) where those number are sorted (small to large, top to bottom). Each comparator (connection between the lines) swaps the numbers on its two lines if they are not in order, otherwise it does nothing. This network has three layers and five comparators, and is the minimal four-input sorting network. Minimal networks are generally not known for large input sizes. Their design space is deceptive which makes network minimization a challenging optimization problem.}
\label{figure:1}       
\end{figure}

\subsection{Sorting Networks}
\label{subsec:5}
A sorting network of \emph{n} inputs is a fixed layout of comparison-exchange operations (comparators) that sorts all inputs of size \emph{n} (Fig.\ref{figure:1}). Since the same layout can sort any input, it represents an oblivious or data-independent sorting algorithm, that is, the layout of comparisons does not depend on the input data. The resulting fixed communication pattern makes sorting networks desirable in parallel implementations of sorting, such as those in graphics processing units, multi-processor computers, and switching networks \cite{ref-1, ref-16, ref-31}. Beyond validity, the main goal in

designing sorting networks is to minimize the number of layers, because it determines how many steps are required in a parallel implementation. A tertiary goal is to minimize the total number of comparators in the networks. Designing such minimal sorting networks is a challenging optimization problem that has been the subject of active research since the 1950s \cite{ref-17}. Although the space of possible networks is infinite, it is relatively easy to test whether a particular network is correct: If it sorts all combinations of zeros and ones correctly, it will sort all inputs correctly \cite{ref-17}.

Many of the recent advances in sorting network design are due to evolutionary methods \cite{ref-32}. However, it is still a challenging problem, even for the most powerful evolutionary methods, because it is highly deceptive: Improving upon a current design may require temporarily growing the network, or sorting fewer inputs correctly. Sorting networks are therefore a good domain for testing the power of evolutionary algorithms.

\begin{figure}
\begin{center}
\includegraphics[scale=.20]{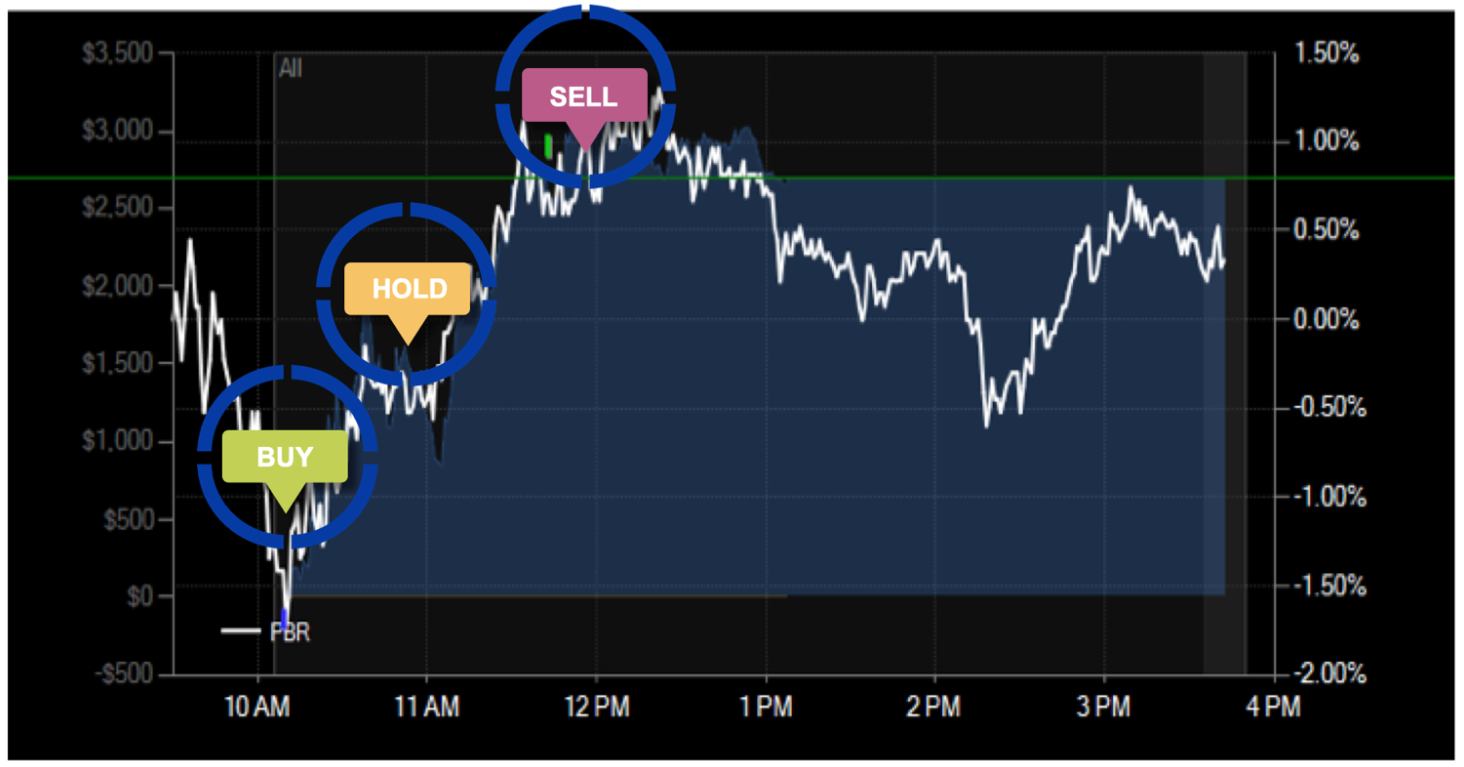}
\end{center}
\caption{Stock Trading Agent. The agent observes the time series of stock prices and makes live decisions about whether to buy, hold, or sell a particular stock. The signal is noisy and prone to overfitting; generalization to unseen data is the main challenge in this domain.}
\label{figure:2}
\end{figure}

\subsection{Stock Trading}
\label{subsec:6}
Stock trading is a natural multi-objective domain where return and risk must be balanced \cite{ref-36, ref-37}. Candidate solutions, i.e. trading agents, can be represented in several ways. Rule-based strategies, sequence modeling with neural networks and LSTMs (Long Short-Term Memory), and symbolic regression using Genetic Programming or Grammatical Evolution are common approaches \cite{ref-38, ref-39}. Frequency of trade, fundamental versus technical indicators, choice of trading instruments, transaction costs, and vocabulary of order types are crucial design decisions in building such agents.

The goal is to extract patterns from historical time-series data on stock prices and utilize those patterns to make optimal trading decisions, i.e. whether to buy, hold, or sell particular stocks (Fig.\ref{figure:2}) \cite{ref-40, ref-41}. The main challenge is to trade in a manner that generalizes to previously unseen situations in live trading. Some general methods like training data interleaving can be used to increase generalization \cite{ref-42}, but their effectiveness might not be enough due to the low signal to noise ratio which is the main source of deception in this domain. The data is extremely noisy and prone to overfitting, and methods that discover more robust decisions are needed.

\section{Methods}
\label{sec:3}
In this section, the genetic representation, the single and multi-objective optimization approaches, the composite objective method, the novelty-based selection method, and the novelty pulsation method are described, using the sorting network domain as an example. These methods were applied to stock trading in an analogous manner.

\subsection{Representation}
\label{subsec:7}
In order to apply various evolutionary optimization techniques to the sorting network problem, a general structured representation was developed. Sorting networks of \(n\) line can be seen as a sequence of two-leg comparators where each leg is connected to a different input line and the first leg is connected to a higher line than the second:\(\{\left( f_{1},\ s_{1} \right),\left( f_{2},\ s_{2} \right),\left( f_{3},\ s_{3} \right),\ldots,\left( f_{c},\ s_{c} \right)\}\).

The number of layers can be determined from such a sequence by grouping successive comparators together into a layer until the next comparator adds a second connection to one of the lines in the same layer. With this representation, mutation and crossover operators amount to adding and removing a comparator, swapping two comparators, and crossing over the comparator sequences of two parents at a single point.

Domain-specific techniques such as mathematically designing the prefix layers \cite{ref-3, ref-4} or utilizing certain symmetries \cite{ref-32} were not used.

\subsection{Single-Objective Approach}
\label{subsec:8}
Correctness is part of the definition of a sorting network: Even if a network mishandles only one sample, it will not be useful. The number of layers can be considered the most important size objective because it determines the efficiency of a parallel implementation. A hierarchical composite objective can therefore be defined as:
\begin{equation}
\label{eq-2}
\\{SingleFitness}\left( m,\ l,\ c \right) = 10000\ m + 100\ l + c\;
\end{equation}
Where \textit{m}, \textit{l}, and \textit{c} are the number of mistakes (unsorted samples), number of layers, and number of comparators, respectively.

In the experiments in this paper, the solutions will be limited to less than one hundred layers and comparators, and therefore, the fitness will be completely hierarchical (i.e. there is no folding).

\subsection{Multi-Objective Approach}
\label{subsec:9}
In the multi-objective approach, the same dimensions, i.e. the number of mistakes, layers, and comparators \(m,\ l,\ c,\) are used as three separate objectives. They are optimized by the NSGA-II algorithm \cite{ref-8} with selection percentage set to 10\%. Indeed, this approach may discover solutions with just a single layer, or a single comparator, since they qualify for the Pareto front. Therefore, diversity is increased compared to the single-objective method, but this diversity is not necessarily helpful.

\subsection{Composite Multi-Objective Approach}
\label{subsec:10}
In order to construct composite axes, each objective is augmented with sensitivity to the other objectives:
\begin{equation}
\label{eq-3}
\\{Composite}_{1}\left( m,\ l,\ c \right) = 10000\ m + 100\ l + c\;
\end{equation}
\begin{equation}
\label{eq-4}
\\{Composite}_{2}\left( m,\ l \right) = \alpha_{1}m + \alpha_{2}l\;
\end{equation}
\begin{equation}
\label{eq-5}
\\{Composite}_{3}\left( m,\ c \right) = \alpha_{3}m + \alpha_{4}c\;
\end{equation}
The primary composite objective (Equation~3), which will replace the mistake axis, is the same hierarchical fitness used in the single-objective approach. It discourages evolution from constructing correct networks that are extremely large. The second objective (Equation~4), with \(\alpha_{2} = 10,\) primarily encourages evolution to look for solutions with a small number of layers. A much smaller cost of mistakes, with \(\alpha_{1} = 1,\ \)helps prevent useless single-layer networks from appearing in the Pareto front. Similarly, the third objective (Equation~5), with \(\alpha_{3} = 1\) and \(\alpha_{4} = 10,\) applies the same principle to the number of comparators.

The values for \(\alpha_{1},\) \(\alpha_{2},\alpha_{3},\ \)and \(\alpha_{4}\) were found to work well in this application, but the approach was found not to be very sensitive to them; A broad range will work as long as they establish a primacy relationship between the objectives.

It might seem like we are adding several hyper-parameters which need to be tuned, but we can estimate them in each domain by picking values that push away trivial or useless solution off the Pareto front.

\subsection{Novelty Selection Method}
\label{subsec:11}
In order to measure how novel the solutions are it is first necessary
to characterize their behavior. While such a characterization can be
done in many ways, a concise and computationally efficient approach is
to count how many swaps took place on each line in sorting all
possible zero-one combinations during the validity check. Such a
characterization is a vector that has the same size as the problem,
making the distance calculations fast. It also represents the true
behavior of the network: Even if two networks sort the same input
cases correctly, they may do it in different ways, and the
characterization is likely to capture that difference. Given this
behavior characterization, novelty of a solution is measured by the
sum of pairwise distances of its behavior vector to those of all the
other individuals in the selection pool:

\begin{equation}
\label{eq-6}
\\{NoveltyScore}\left( x_{i} \right) = \sum_{j = 1}^{n}d(b\left( x_{i}
\right),\ b\left( x_{j} \right))\; .
\end{equation}

\begin{figure}
\begin{center}
\includegraphics[scale=.24]{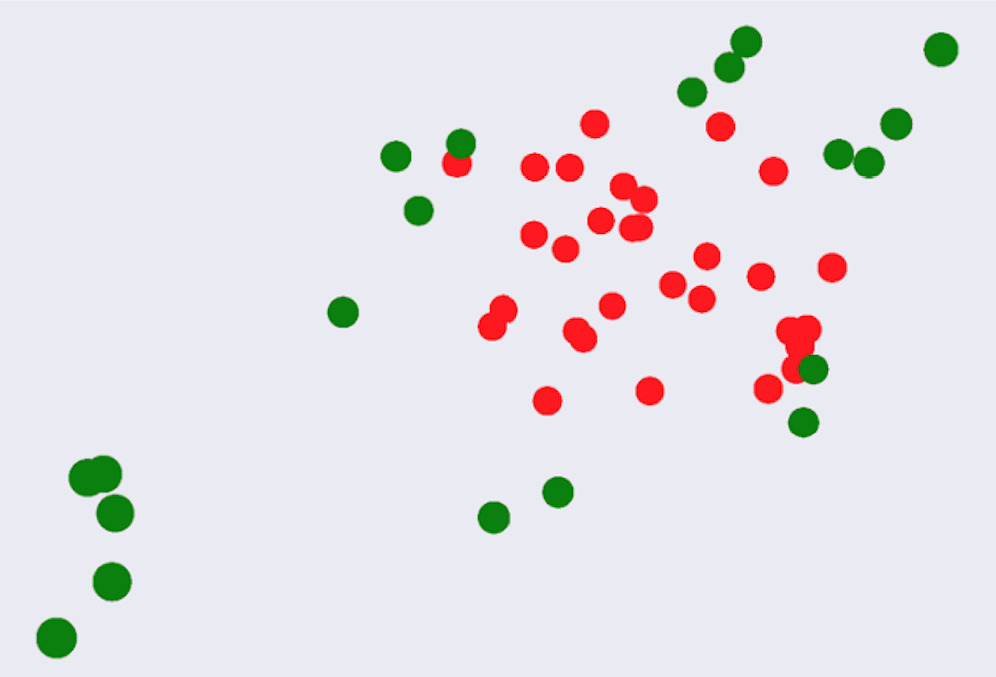}
\end{center}
\caption{The first phase of novelty selection is to select the solutions (marked green) with the highest Novelty Score (Equation~\ref{eq-6}).}
\label{figure:7}
\end{figure}

The selection method also has a parameter called \emph{selection multiplier} (e.g. set to 2 in these experiments), varying between one and the inverse of the elite fraction (e.g. 1/10, i.e. 10\%) used in the NSGA-II multi-objective optimization method. The original selection percentage is multiplied by the selection multiplier to form a broader selection pool. That pool is sorted according to novelty, and the top fraction representing the original selection percentage is used for selection. This way, good solutions that are more novel are included in the pool.
Fig.\ref{figure:7} shows an example result of applying
Equation~\ref{eq-6}.

\begin{figure}
\begin{center}
\includegraphics[scale=.24]{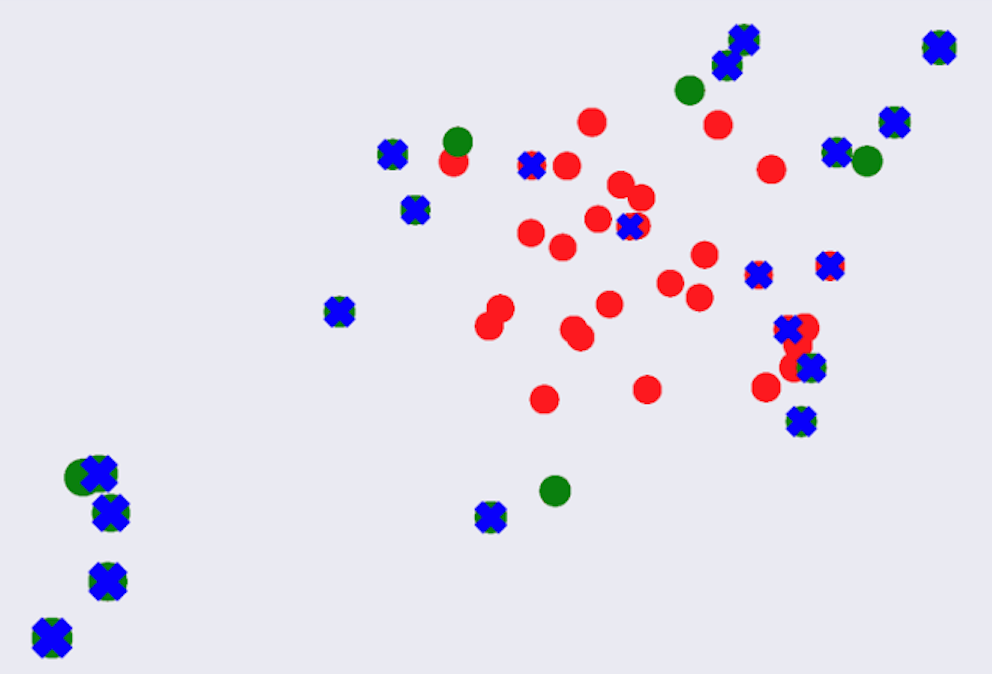}
\end{center}
\caption{Phase two of novelty selection eliminates the closest pairs
  of the green candidates in order to get better overall coverage
  (blue candidates). The result is a healthy mixture of high-fitness
  candidates and novel ones (Equation~\ref{eq-7}).} 
\label{figure:8}
\end{figure}

One potential issue is that a cluster of solutions far from the rest may end up having high novelty scores while only one is good enough to keep. Therefore, after the top fraction is selected, the rest of the sorted solutions are added to the selection pool one by one, replacing the solution with the lowest minimum novelty, defined as

\begin{equation}
\label{eq-7}
\\{MinimumNovelty}\left( x_{i} \right) = \min_{1\leq j \leq
  n;\ j\  \neq \ i}{d(b\left( x_{i} \right),\ b\left( x_{j}
  \right))}\; .
\end{equation}

Note that this method allows tuning novelty selection continuously
between two extremes: By setting \emph{selection multiplier} to one,
the method reduces to the original multi-objective method (i.e.\ only
the elite fraction ends up in the final elitist pool), and setting it
to the inverse of the elite fraction reduces it to pure novelty search
(i.e. the whole population, sorted by novelty, is the selection
pool). In practice, low and midrange values for the multiplier work
well, including the value 2 used in these experiments.
Fig.\ref{figure:8} shows an example result of applying
Equation~\ref{eq-7}.  The entire novelty-selection algorithm is
summarized in Fig.~\ref{fg:noveltyselectionalgorithm}.

\begin{figure}
\begin{center}
\begin{svgraybox}
\setlength{\parskip}{1pt}
\setlength{\parindent}{5pt}
\begin{enumerate}
\item
\emph{using a selection method (e.g. NSGA-II) pick \textbf{selection
    multiplier} times as many elitist candidates than usual}\\
\item
\emph{sort them in descending order according to their NoveltyScore (Equation~\ref{eq-6})}\\
\item
\emph{move the usual number of elitist candidates from the top of the list to the result set}\\
\item
\emph{for all remaining candidates in the sorted list:}
\begin{enumerate}
\item
\emph{add the candidate to result set}
\item
\emph{remove the candidate with the lowest MinimumNovelty (Equation~\ref{eq-7})}
\item
\emph{(using a fitness measure as the tie breaker)}\\
\end{enumerate}
\item
\emph{return the resulting set as the elite set}
\end{enumerate}
\end{svgraybox}
\end{center}
\caption{The Novelty Selection Algorithm}
\label{fg:noveltyselectionalgorithm}
\end{figure}

\begin{figure}
\begin{center}
\includegraphics[scale=.22]{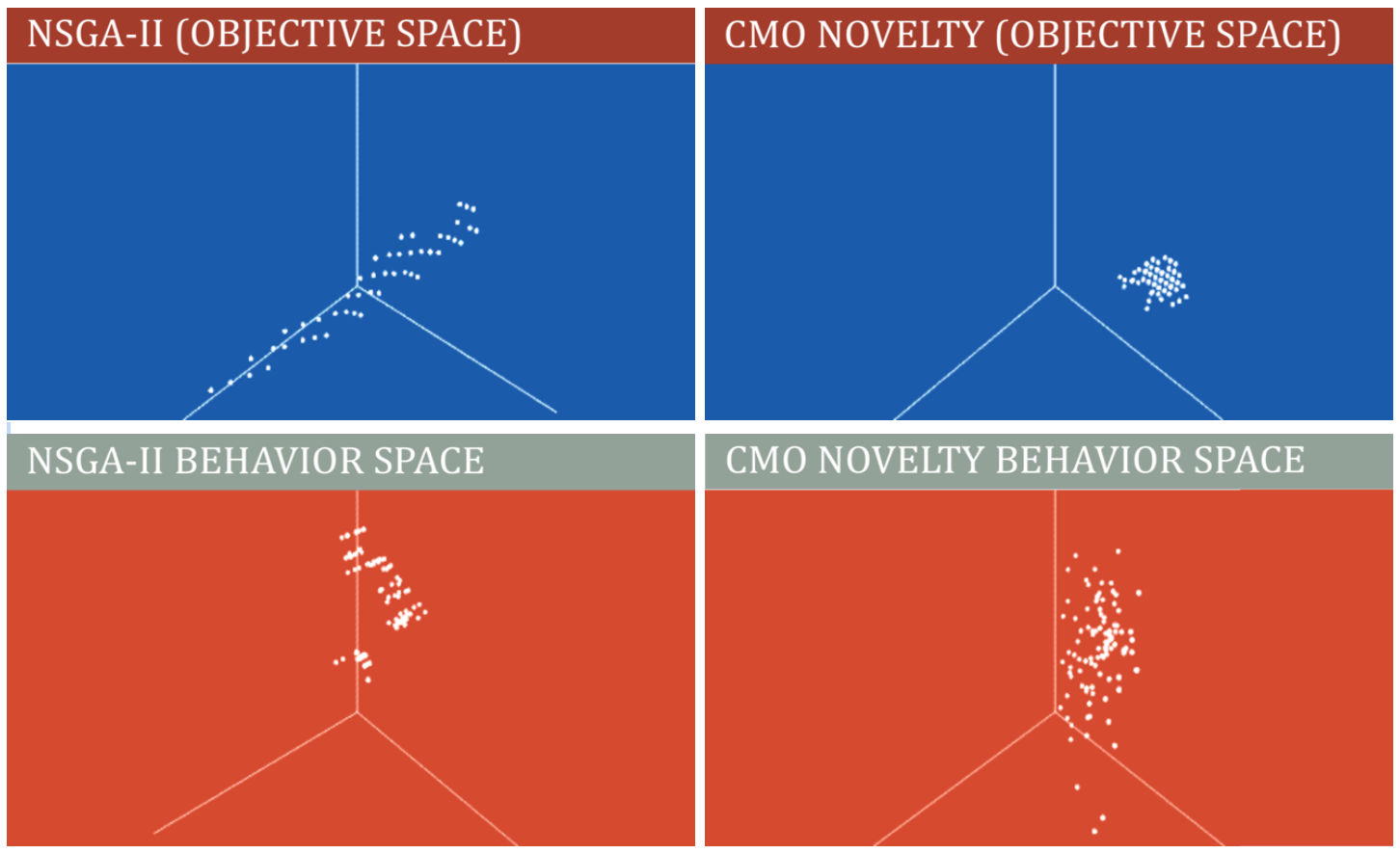}
\end{center}
\caption{An example demonstrating how novelty selection (right column)
  creates better coverage of the behavior space (bottom row) than
  NSGA-II (left column) despite being more focused in the objective
  space (top row).}
\label{figure:9}
\end{figure}

To visualize this process, Fig.\ref{figure:9} contrasts the difference
between diversity that multi-objective method (e.g. NSGA-II) creates
(left-side) and diversity that novelty search creates (right-side). In
the objective space (top), novelty looks more focused and less
diverse, but in the behavior space (bottom) it is much more
diverse. This type of diversity enables the method to escape
deception and find novel solutions, such as the state of
the art in sorting networks.

\subsection{Novelty Pulsation Method}
\label{subsec:12}
Parent selection is a crucial step in an evolutionary algorithm. In
almost all such algorithms, whatever method is used remains unchanged
during the evolutionary run. However, 
when a problem is deceptive or prone to over-fitting, changing the
selection method periodically may make the algorithm more robust. It
can be used to alternate the search between exploration and
exploitation, and thus find a proper balance between them.

In Composite Novelty Pulsation, novelty selection is switched on and off after a certain number of generations. As in delta-coding and burst mutation, once good solutions are found, they are used as a starting point for exploration. Once exploration has generated sufficient diversity, local optimization is performed to find the best possible versions of these diverse points. These two phases leverage each other, which results in faster convergence and more reliable solutions.

\begin{figure}
\begin{center}
\includegraphics[scale=.22]{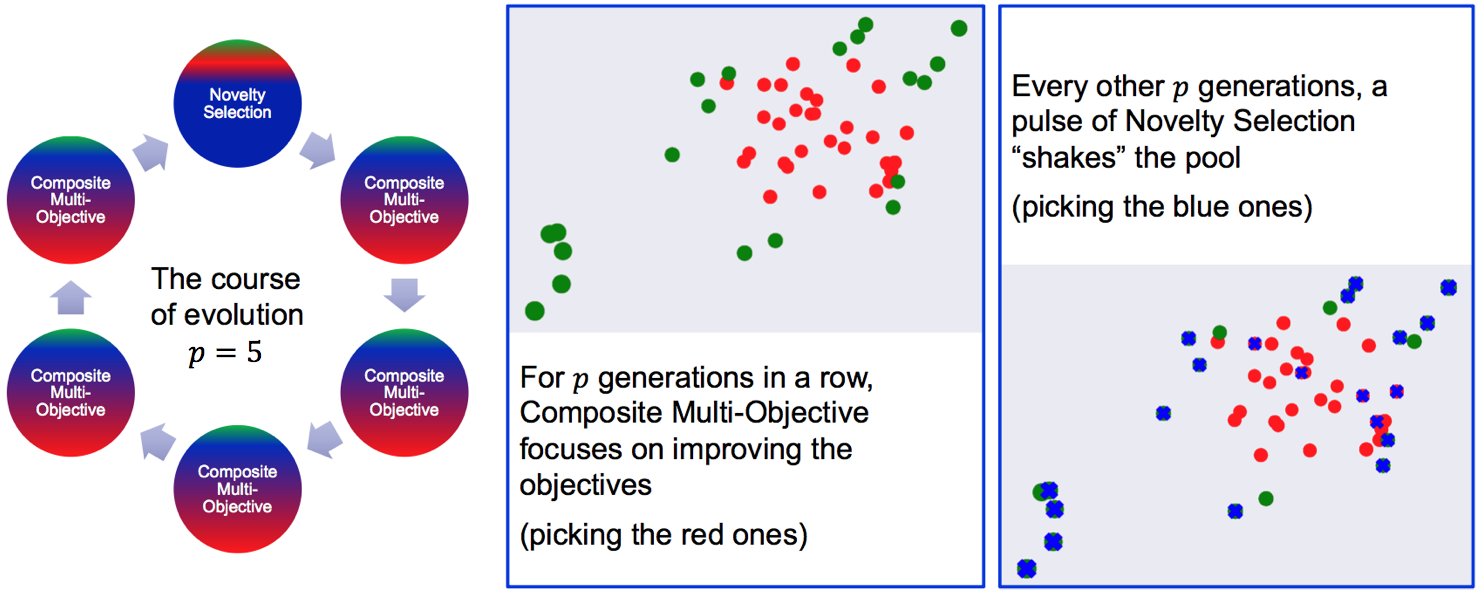}
\end{center}
\caption{Visulization of how novelty pulsation process alternates between composite multi-objective selection and novelty selection.}
\label{figure:10}
\end{figure}

Composite Novelty Pulsation adds a new hyper-parameter, \(P\),
denoting the number of generations before switching novelty selection
on and off. Preliminary experiments showed that \(P = 5\) works well
in both sorting network and stock trading domains; however, in
principle it is possible to tune this parameter to fit the domain.
Fig.\ref{figure:10} shows the novelty pulsation process schematic.

\section{Experiment}
\label{sec:4}
Previous work in the sorting networks domain demonstrated that composite novelty can match the minimal known networks up to 18 input line with reasonable computational resources \cite{ref-31, ref-32}. The goal of the sorting network experiments was to achieve the same result faster, i.e. with fewer resources. The experiments were therefore standardized to a single machine (a multi-core desktop).

In the stock market trading domain, the experiments compared generalization by measuring the correlation between seen and unseen data.

\subsection{Experimental Setup}
\label{subsec:13}
Experiments in previous paper \cite{ref-31} already demonstrated that
the composite novelty method performs statistically significantly
better in the sorting network discovery task than the other methods
discussed above. Therefore, this paper focuses on comparing the novelty
pulsation method to its predecessor, i.e.\ the composite novelty method.

In the sorting networks domain, experiments were run with the
following parameters:

\begin{itemize}
\item Eleven network sizes, 8 through 18;
\item Ten runs for each configuration (220 runs in total);
\item 10\% parent selection rate;
\item Population size of 1000 for composite novelty selection and 100 for novelty pulsation. These settings were found to be appropriate for each method experimentally.
\end{itemize}

In the trading domain, experiments were run with the following parameters:

\begin{itemize}
\item Ten runs on five years of historical data;
\item Population size of 500;
\item 100 generations;
\item 10\% parent selection rate;
\item Performance of the 10 best individuals from each run compared on the subsequent year of historical data, withheld from all runs.
\end{itemize}

\begin{figure}
\begin{center}
\includegraphics[scale=.27]{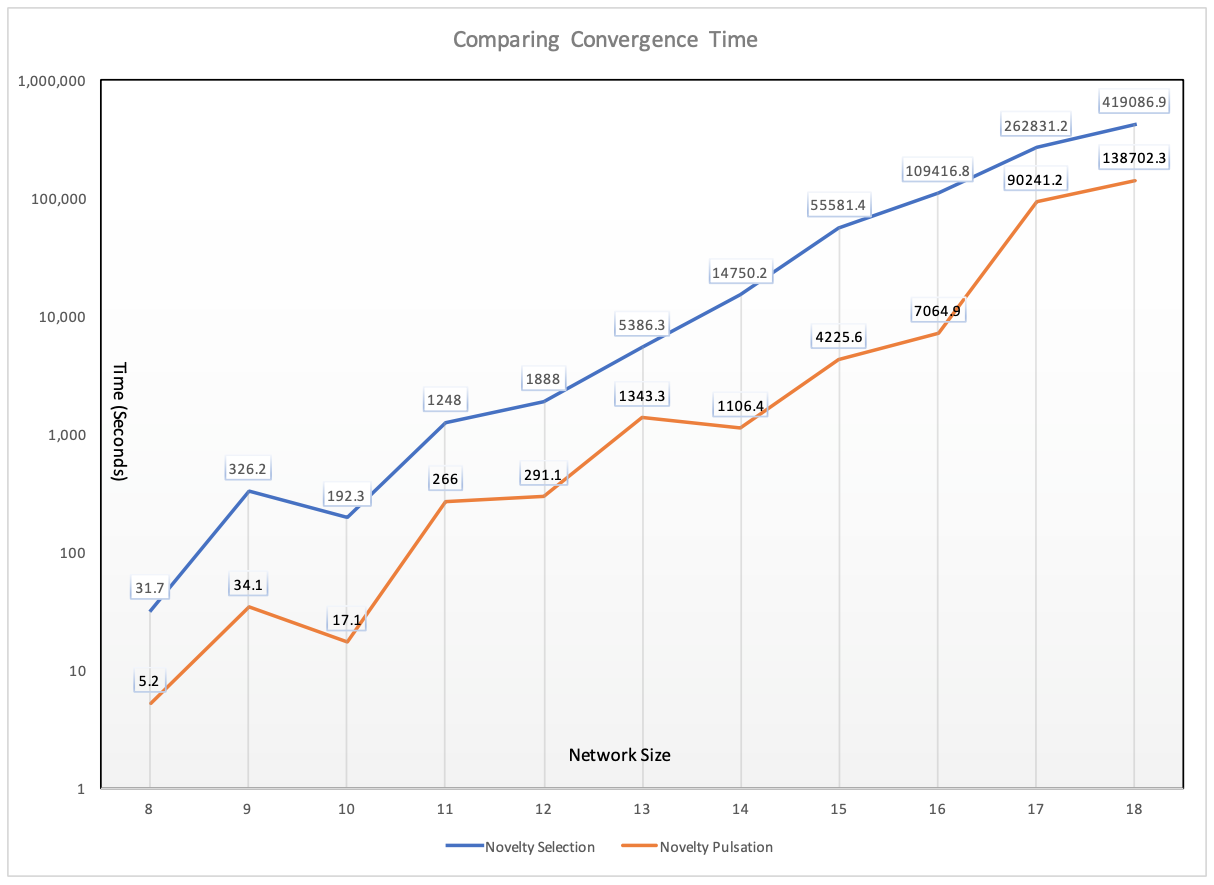}
\end{center}
\caption{The average runtime needed to converge to the state of the
  art on networks with different sizes. Novelty pulsation converges significantly faster at all sizes, demonstrating improved balance of exploration and exploitation.}
\label{figure:3}
\end{figure}

\subsection{Sorting Networks Results}
\label{subsec:14}
Convergence time of the two methods to minimal solutions for different network sizes is shown in Fig.\ref{figure:3}. Novelty pulsation shows an order of magnitude faster convergence across the board. All runs resulted in state-of-the-art sorting networks.

An interesting observation is that sorting networks with an even
number of lines take proportionately less time to find the
state-of-the-art solution than those with odd numbers of lines. This
result is likely due to symmetrical characteristics of even-numbered
problems. Some methods \cite{ref-32} exploit this very symmetry in
order to find state-of-the-art solutions and break previous records,
but this domain-specific information was not used in the
implementation in this paper. The fact that the method achieves the
state-of-the-art results and even breaks one world record (as
described in the Appendix) without exploiting domain specific
characteristics is itself a significant result.

\subsection{Stock Trading Results}
\label{subsec:15}
\begin{figure}
\begin{center}
\includegraphics[scale=.25]{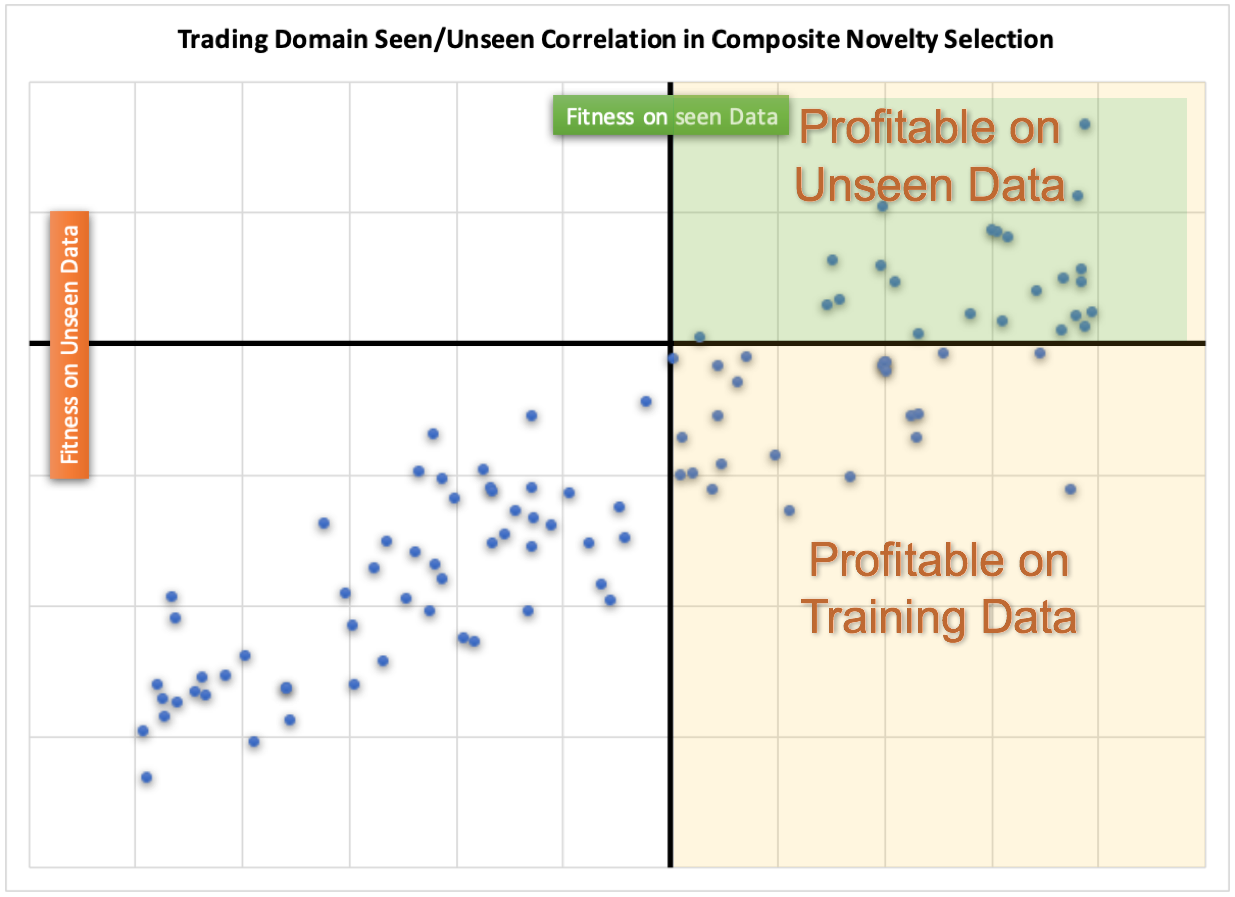}
\end{center}
\caption{Generalization from seen to unseen data with the
  Composite Novelty method. The fitness on seen data is on $x$ and
  unseen is on $y$. The correlation is 0.69, which is enough
  to trade but could be improved. Candidates to the right of the
  vertical line are profitable on seen data, and candidates above
  the horizontal line are profitable on unseen data.} 
\label{figure:4}
\end{figure}

\begin{figure}
\begin{center}
\includegraphics[scale=.25]{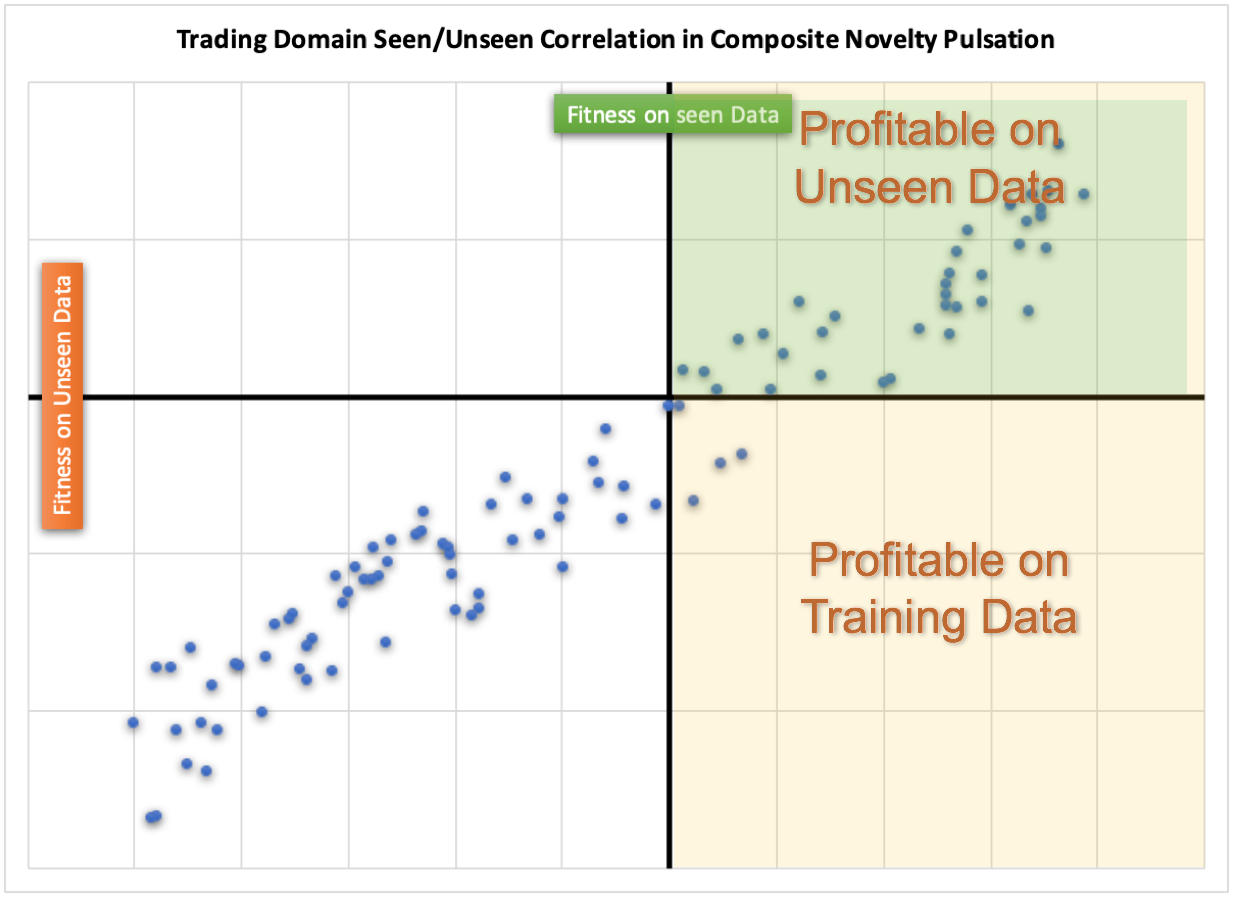}
\end{center}
\caption{Generalization from seen to unseen data with
  Composite Novelty Pulsation method. The correlation is 0.89, which
  results in significantly improved profitability in live trading. It
  is also notable that the majority of profitable genes on training
  data are also profitable on unseen data.}
\label{figure:5}
\end{figure}

Figures~\ref{figure:4} and~\ref{figure:5} illustrate generalization of
the composite novelty selection and novelty pulsation methods,
respectively. Points in Fig.\ref{figure:5} are noticeably closer to a
diagonal line, which means that better training fitness resulted in
better testing fitness, i.e. higher correlation and better
generalization. Numerically, the seen-to-unseen correlation for the
composite novelty method is 0.69, while for composite novelty
pulsation, it is 0.86. The ratio of the number of profitable
candidates on unseen data and training data is also better, suggesting
that underfitting is unlikely. In practice, these differences are
significant, translating to much improved profitability in live
trading.

\section{Discussion and Future Work}
\label{sec:5}
The results in both sorting network and stock trading domains support the anticipated advantages of the composite novelty pulsation approach. The secondary objectives diversify the search, composite objectives focus it on most useful areas, and pulses of novelty selection allow for both accurate optimization and thorough exploration of those areas. These methods are general and robust: they can be readily implemented in standard multi-objective search such as NSGA-II and used in combination with many other techniques already developed to improve evolutionary multi-objective optimization.

The sorting network experiments were designed to demonstrate the improvement provided by novelty pulsation over the previous state of the art. Indeed, it found the best known solutions significantly faster. One compelling direction of future work is to use it to optimize sorting networks systematically, with domain-specific techniques integrated into the search, and with significantly more computing power, including distributed evolution \cite{ref-14}. It is likely that given such power, many new minimal networks can be discovered, for networks with even larger number of input lines.

The stock trading experiments was designed to demonstrate that the approach makes a difference in real-world problems. The main challenge in trading is generalization to unseen data, and indeed in this respect novelty pulsation improved generalization significantly.

The method can also be applied in many other domains, in particular those that are deceptive and have natural secondary objectives. For instance, various game strategies from board to video games can be cast in this form, where winning is accompanied by different dimensions of the score. Solutions for many design problems, such as 3D printed objects, need to satisfy a set of functional requirements, but also maximize strength and minimize material. Effective control of robotic systems need to accomplish a goal while minimize energy and wear and tear. Thus, many applications should be amenable to the composite novelty pulsation approach.

Another direction is to extend the method further into discovering effective collections of solutions. For instance, ensembling is a good approach for increasing the performance of machine learning systems. Usually the ensemble is formed from solutions with different initialization or training, with no mechanism to ensure that their differences are useful. In composite novelty pulsation, the Pareto front consists of a diverse set of solutions that span the area of useful tradeoffs. Such collections should make for a powerful ensemble, extending the applicability of the approach further.

\section{Conclusion}
\label{sec:6}
The composite novelty pulsation method is a promising extension of the composite novelty approach to deceptive problems. Composite objectives focus the search on the most useful tradeoffs (better exploitation), while novelty selection allows escaping deceptive areas (better exploration). Novelty pulsation balances between the exploration and exploitation, finding solutions faster and finding solutions that generalize better. These principles were demonstrated in this paper in the highly deceptive problem of minimizing sorting networks and in the highly noisy domain of stock market trading. Composite novelty pulsation is a general method that can be combined with other advances in population-based search, thus increasing the power and applicability of evolutionary multi-objective optimization.

\section*{Appendix}
\label{sec:app}
The graph of the new world record for 20-line sorting network, which moved the previous record of 92 comparators also discovered by evolution \cite{ref-32} down to 91.
\begin{figure}
\begin{center}
\includegraphics[scale=.23]{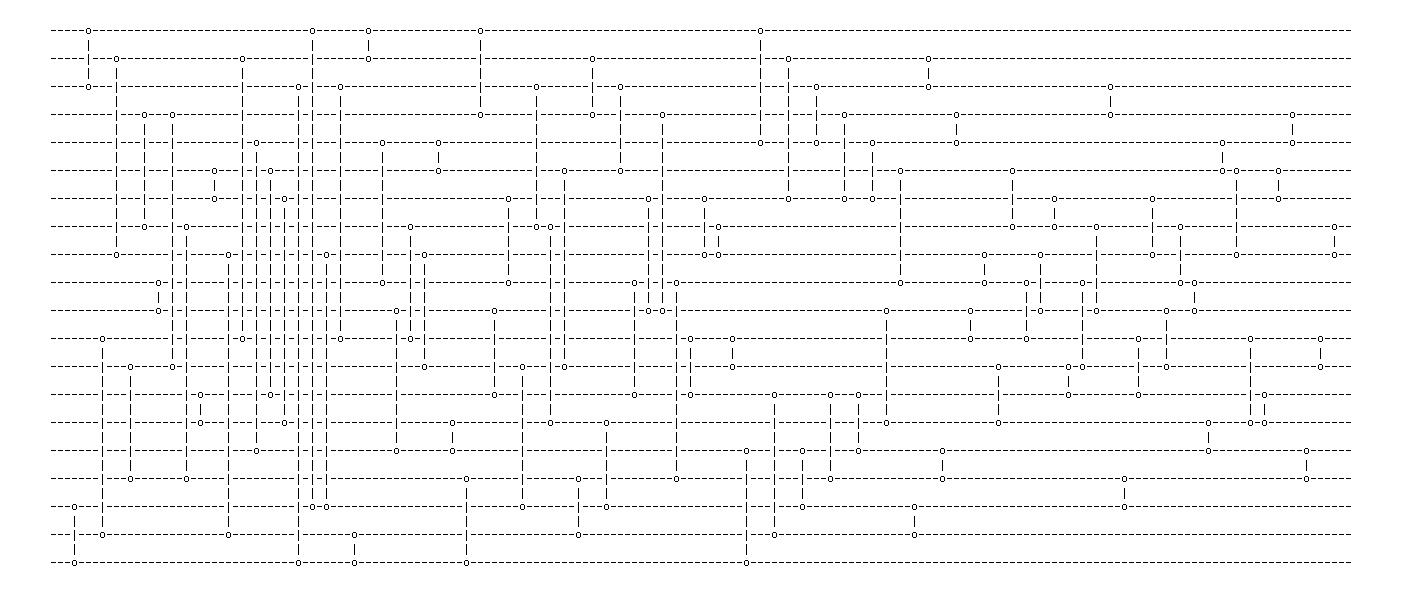}
\end{center}
\caption{The new 20-line sorting network with 91 comparators, discovered by Novelty Pulsation.}
\label{figure:6}
\end{figure}

\begin{svgraybox}
One of the nice properties of Novelty Pulsation Method is the ability to converge with a very small pool size (like only 30 individuals in case of sorting networks). However, it still took almost two months to break the world record on the 20-line network running on a single machine (Fig.\ref{figure:6}). Interestingly, even if it takes the same number of generations for the other methods to get there with a normal pool size of a thousand, those runs will take almost five years to converge! 
\end{svgraybox}

%
%
%

\end{document}